\documentclass{article}

\usepackage{microtype}
\usepackage{graphicx}
\usepackage{booktabs}

\usepackage{hyperref}

\usepackage[preprint]{icml2026}
\makeatletter
\renewcommand{\Notice@String}{Accepted at the Workshop on Forecasting as a New Frontier of Intelligence, ICML 2026, Seoul, South Korea.}
\makeatother

\usepackage{amsmath,amssymb}
\usepackage{multirow}

\graphicspath{{figures/}}

\icmltitlerunning{Do TSFM Benchmarks Hide Regime-Dependent Failures?}

\begin{document}

\twocolumn[
  \icmltitle{Do Time Series Foundation Model Benchmarks Hide Regime-Dependent Failures?\\Evidence from Traffic Speed Forecasting}

  \begin{icmlauthorlist}
    \icmlauthor{Yingshuo Wang}{ucb}
    \icmlauthor{Xian Sun}{duke}
    \icmlauthor{Lingdong Kong}{nus}
    \icmlauthor{Wei Gao}{neu}
    \icmlauthor{Yanhang Li}{neu}
    \icmlauthor{Zhichao Fan}{uiuc}
    \icmlauthor{Zexin Zhuang}{smu}
  \end{icmlauthorlist}

  \icmlaffiliation{ucb}{University of California, Berkeley, CA, USA}
  \icmlaffiliation{duke}{Duke University, Durham, NC, USA}
  \icmlaffiliation{nus}{National University of Singapore, Singapore}
  \icmlaffiliation{neu}{Northeastern University, Boston, MA, USA}
  \icmlaffiliation{uiuc}{University of Illinois Urbana-Champaign, IL, USA}
  \icmlaffiliation{smu}{Southern Methodist University, Dallas, TX, USA}

  \icmlcorrespondingauthor{Yingshuo Wang}{yingshuow@berkeley.edu}

  \icmlkeywords{time series foundation models, probabilistic forecasting, regime switching, benchmark evaluation}

  \vskip 0.3in
]

\printAffiliationsAndNotice{}

\begin{abstract}
Standard benchmarks evaluate time series foundation models (TSFMs) using aggregate metrics, but these can mask severe failures in critical operating regimes.
We introduce \emph{regime-stratified evaluation} and apply it to three TSFMs on two standard traffic speed benchmarks.
Traffic exhibits abrupt regime switching between free-flow and congested states, producing bimodal speed distributions during transitions.
When we stratify by traffic regime, both accuracy and prediction-interval coverage degrade sharply during transitions: transition-regime MAE reaches 11\,mph (versus 3\,mph overall), and empirical coverage of 90\% prediction intervals drops as low as 55\%.
These failures are invisible in aggregate metrics because free-flow observations dominate the sample.
A simple historical conditional baseline (sampling from per-sensor training distributions) achieves better transition coverage than any TSFM, but has far worse overall accuracy.
We propose \emph{bimodal mixture augmentation} (BMA), a post-hoc method that combines TSFM forecasts with historical distributional knowledge, approaching the historical baseline's transition coverage while preserving the TSFM's accuracy.
Our results suggest that TSFM benchmarks should incorporate regime-aware evaluation to surface failures that aggregate metrics hide.
\end{abstract}

\section{Introduction}
\label{sec:intro}

Traffic speed forecasting matters most during transitions between free-flow and congestion.
In stable conditions, real-time data suffices; during onset or dissipation of congestion, traffic can break down or recover, and systems that consume future speeds (estimated times of arrival (ETAs), freight windows, emergency dispatch) need forecasts with reliable uncertainty.
This is where forecasting earns its value.

Time series foundation models (TSFMs) are positioned as general-purpose probabilistic forecasters \citep{ansari2024chronos,woo2024moirai}.
TSFM-Bench \citep{li2025tsfmbench} and GIFT-Eval \citep{aksu2025gifteval} include METR-LA (207 loop detectors, Los Angeles) and PEMS-BAY (325 detectors, San Francisco Bay Area) \citep{li2018dcrnn}, the standard benchmarks used by DCRNN \citep{li2018dcrnn}, STGCN \citep{yu2018stgcn}, and subsequent graph neural network models.
These leaderboards stratify by domain and frequency but not by operating regime within a domain.
A model can appear accurate on average while failing in the regimes that matter most.
\citet{adler2025calibration} studied TSFM prediction-interval coverage on six general datasets and found foundation models ``consistently better calibrated,'' but did not include high-frequency traffic data.
Traffic at 5-minute resolution exhibits \emph{abrupt regime switching} between free-flow ($\sim$65\,mph) and congested ($\sim$10--20\,mph) states \citep{greenshields1935}, driven by sharp capacity thresholds.
This is a property of traffic physics, not of any particular sensor.
During transitions, the true future speed distribution is bimodal (Figure~\ref{fig:distributions}): speed will either remain high or drop sharply.
TSFMs produce unimodal intervals centered between the modes ($\sim$30--45\,mph), a range that is transient and rarely sustained.

\paragraph{Related work.}
Conformal methods have been extended to non-stationary series: \citet{gibbs2021aci} introduced adaptive conformal inference (ACI); recent work addresses regime switching \citep{regimeswitchingaci2025} and correlated series \citep{cini2025relational}.
All adjust interval \emph{width} but cannot correct distributional \emph{shape}.
On the traffic side, \citet{wu2023quantraffic} and \citet{zheng2025probtraf} proposed uncertainty quantification for deep traffic models, but train domain-specific architectures from scratch and do not evaluate zero-shot TSFMs.
Widening a unimodal interval centered on 40\,mph still misses outcomes near 15 or 65\,mph; a different approach is needed.

We present the first \emph{regime-stratified} evaluation of TSFMs on standard traffic speed benchmarks.
Our contributions are:
\textbf{(1)}~We show that aggregate metrics mask severe regime-dependent failures in \emph{both} accuracy and prediction-interval coverage during traffic regime transitions.
\textbf{(2)}~We show that a simple historical conditional baseline achieves better transition coverage than any TSFM but worse overall accuracy, revealing complementary strengths.
\textbf{(3)}~We propose \emph{bimodal mixture augmentation} (BMA), which combines TSFM forecasts with historical distributional knowledge, improving transition coverage by 3--19\,pp without retraining.

\section{Experimental Setup}
\label{sec:setup}

\paragraph{Datasets.}
We use METR-LA (207 sensors, Los Angeles freeways, Mar--Jun 2012) and PEMS-BAY (325 sensors, San Francisco Bay Area, Jan--May 2017) \citep{li2018dcrnn}, the standard benchmarks for spatio-temporal traffic forecasting.
Both record speed at 5-minute intervals.
We use the standard 70/10/20 train/validation/test split.

\paragraph{Sensor selection.}
To focus on the regimes where coverage failures matter most, we rank sensors by their training-set congestion frequency (fraction of readings below 25\,mph) and form a pool of the top~50 per dataset.
Each random seed then subsamples 30 sensors from this pool, introducing variability across runs.
This uses only training data, avoiding test-set leakage.

\paragraph{Models.}
We evaluate three TSFMs in zero-shot mode: \textbf{Chronos-T5-Base}~\citep{ansari2024chronos}, a T5-based encoder-decoder that tokenizes time series values and generates 100 distributional samples; \textbf{Chronos-Bolt-Small}~\citep{ansari2024chronos}, an efficient quantile-based variant from which we draw pseudo-samples via interpolation; and \textbf{Moirai-1.1-R-Base}~\citep{woo2024moirai}, which uses any-variate attention with mixture outputs and generates 100 samples.
Baselines include \textbf{ACI-LR} (linear regression wrapped with adaptive conformal inference \citep{gibbs2021aci}) and a \textbf{historical conditional} baseline that draws 100 samples directly from the per-sensor empirical distribution $P(\text{speed}_{t+h} \mid \text{speed}_t)$ computed from training data, with no forecasting model involved.
The historical conditional tests whether BMA's TSFM anchor adds value beyond the historical lookup it uses.

\paragraph{Evaluation protocol.}
From the pool of 50 congestion-prone sensors, we draw three different random subsets of 30 sensors each (using seeds 42, 43, 44) so that means and standard deviations reflect variability across sensor selections.
For each subset, we evaluate on 50 test windows at horizons $h \in \{3, 6, 12\}$ steps (15, 30, 60 minutes ahead), with 14 hours of context for foundation models.
We report \emph{mean absolute error} (MAE) and \emph{empirical coverage}: the fraction of true observations that fall within the model's 90\% prediction interval.
Differences between coverage rates are reported in percentage points (pp).

\begin{figure}[t]
\centering
\includegraphics[width=\columnwidth]{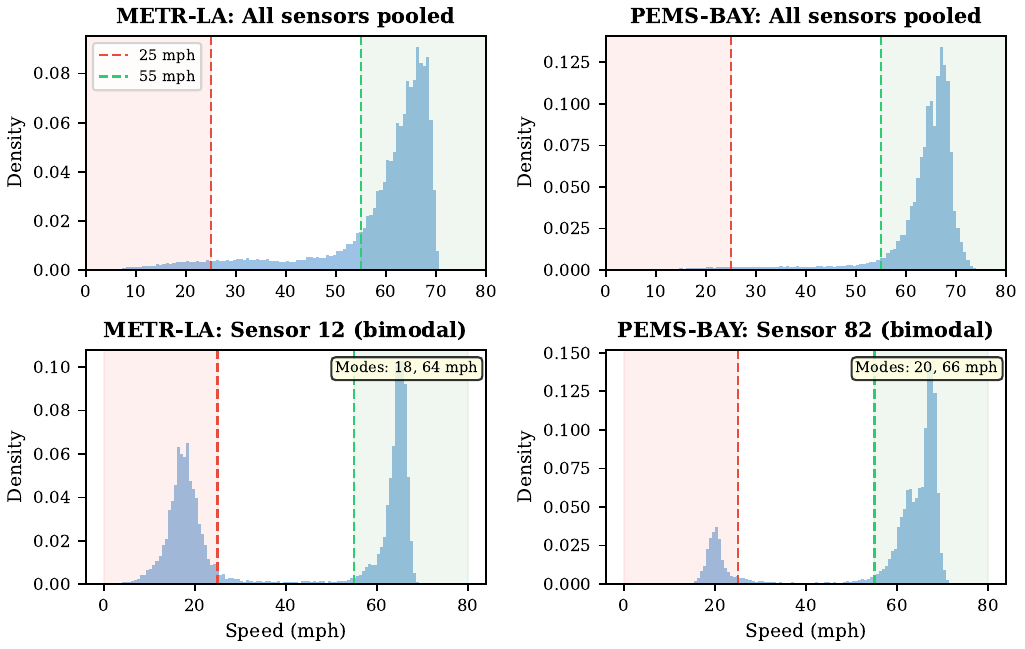}
\caption{Speed distributions from training data. \textbf{Top:} pooled across all sensors, the distribution appears unimodal near 60--65\,mph. \textbf{Bottom:} individual congestion-prone sensors reveal strong bimodality with modes near $\sim$18\,mph and $\sim$64--66\,mph. Red and green lines mark the regime thresholds (25 and 55\,mph); shaded regions indicate the congested (red) and free-flow (green) regimes.}
\label{fig:distributions}
\end{figure}

\paragraph{Regime detection.}
We classify each (window, sensor) target period using Highway Capacity Manual \citep{hcm2022} thresholds: \emph{free-flow} (all steps $>$55\,mph, LOS~A/B), \emph{congested} (all steps $<$25\,mph, LOS~E/F), or \emph{transition} (mixed/intermediate, LOS~C/D).
These thresholds align with the bimodal speed distributions in Figure~\ref{fig:distributions}: individual congestion-prone sensors show modes near $\sim$18 and $\sim$65\,mph with a valley near 30--40\,mph.

\paragraph{Post-hoc correction methods.}
We compare four post-hoc correction strategies, all applied to the cached TSFM samples without retraining:

\emph{Global ACI (G-ACI)}: Adaptive conformal inference \citep{gibbs2021aci} tracks a miscoverage rate $\alpha_t$ updated sequentially: when recent observations fall outside the interval, $\alpha_t$ shrinks, widening future intervals.
Because $\alpha_t$ is shared across regimes, free-flow observations (near-nominal) dilute the transition-regime signal.

\emph{Regime-Conditional ACI (R-ACI)}: Maintains separate $\alpha$ parameters per regime, so transition-regime errors drive wider transition intervals without affecting free-flow.
Both ACI variants adjust interval \emph{width} only and cannot shift probability mass toward a missing mode.

\emph{Bimodal Mixture Augmentation (BMA)}: Our proposed method.
From training data, we precompute per-sensor historical transition probabilities: $P(\text{speed}_{t+h} \in R \mid \text{speed}_t)$ for each regime $R$ and horizon $h$.
At test time, we replace a fraction of the TSFM's 100 forecast samples with draws from this historical conditional distribution, scaled by the transition probability.
This injects the missing mode (e.g., congested speeds $\sim$15\,mph) when historical data indicates a non-trivial chance of regime switching.
The mixing weight $w \in [0.1, 0.5]$ is selected on 10 held-out windows and evaluated on all remaining data.

\emph{BMA + ACI}: BMA fixes the distributional \emph{shape} by injecting the missing mode, but may leave residual over- or undercoverage because the mixing weight $w$ is fixed per configuration.
Global ACI then adjusts the interval \emph{width} sequentially to close any remaining coverage gap.
The two corrections are complementary: shape first, then width.

\section{Results}
\label{sec:results}

\subsection{Aggregate Metrics Hide Regime-Dependent Failures}

Table~\ref{tab:regime_mae} reports MAE stratified by traffic regime at $h$=12.
TSFMs achieve much lower overall MAE than the historical conditional baseline (5.8 vs 12.9\,mph on METR-LA), confirming that the TSFM provides genuine forecasting value.
But during transitions, all methods converge to similar error ($\sim$10--11\,mph) because neither the TSFM nor the historical lookup can predict which direction traffic will move.
Free-flow dominates the sample, so the aggregate is pulled toward the easy regime.

\begin{table}[t]
\caption{MAE (mph) at $h$=12 stratified by traffic regime. TSFMs achieve much lower overall MAE than the historical conditional baseline, but all methods fail similarly during transitions.}
\label{tab:regime_mae}
\vskip 0.05in
\centering
\footnotesize
\setlength{\tabcolsep}{3pt}
\begin{tabular}{@{}l l cccc@{}}
\toprule
& Method & Overall & Free & Trans. & Cong. \\
\midrule
\multirow{4}{*}{\rotatebox[origin=c]{45}{\scriptsize METR-LA}}
& Hist. cond.  & 12.90 & --- & 9.47 & --- \\
& Chronos-T5   & 5.77 & 2.13 & 9.83 & 1.19 \\
& Chronos-Bolt & 6.20 & 2.63 & 9.99 & 2.34 \\
& Moirai       & 5.84 & 2.16 & 9.72 & 1.89 \\
\midrule
\multirow{4}{*}{\rotatebox[origin=c]{45}{\scriptsize P-BAY}}
& Hist. cond.  & 3.11 & --- & 11.05 & --- \\
& Chronos-T5   & 3.07 & 1.35 & 11.04 & 2.70 \\
& Chronos-Bolt & 3.04 & 1.29 & 11.11 & 2.86 \\
& Moirai       & 3.09 & 1.19 & 11.33 & 4.89 \\
\bottomrule
\end{tabular}
\end{table}

\subsection{Calibration Failures Concentrate in Transitions}

\begin{figure}[t]
\centering
\includegraphics[width=\columnwidth]{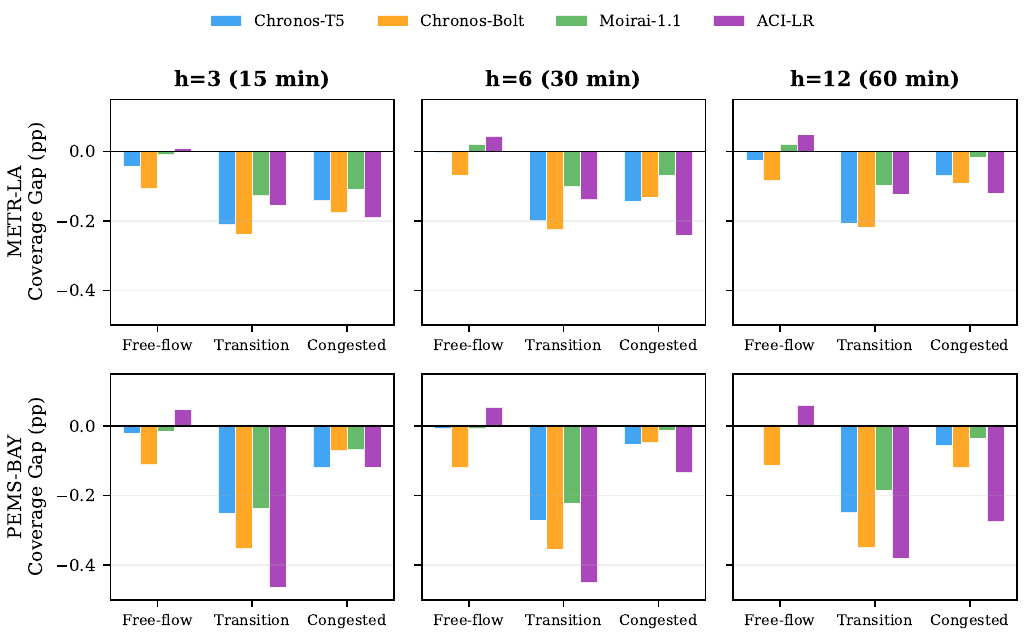}
\caption{Coverage gap by traffic regime at three horizons.
Each bar shows empirical coverage minus the 90\% nominal target; the zero line means the model's intervals achieve exactly 90\% coverage.
Negative bars indicate undercoverage (intervals too narrow).
All models fail during \textbf{transitions}, with gaps reaching $-$35\,pp (Chronos-Bolt) and $-$48\,pp (ACI-LR) on PEMS-BAY.
ACI-LR overcorrects in free-flow (bars above zero) but has the worst transition coverage, illustrating that widening intervals alone cannot fix the shape mismatch.}
\label{fig:regime}
\end{figure}

We construct 90\% prediction intervals for each model (the standard operating point in TSFM benchmarks \citep{adler2025calibration,aksu2025gifteval}), meaning the interval should contain the true speed 90\% of the time, and measure how often it actually does (empirical coverage).
We then stratify by traffic regime.
Figure~\ref{fig:regime} reveals a striking pattern.
During \textbf{free-flow} ($>$55\,mph), TSFMs achieve coverage near the 90\% target (bars near zero).
During \textbf{congestion} ($<$25\,mph), coverage drops moderately.
The \textbf{transition} regime (25--55\,mph) is the worst: Chronos-Bolt achieves only 54.9\% on PEMS-BAY at $h$=12, a 35\,pp gap.
The ACI-LR baseline (linear regression with adaptive conformal inference) is even worse during transitions on PEMS-BAY despite overcovering in free-flow, confirming that width-only adjustment cannot fix this problem.

The transition regime's severe undercoverage matters most when operators need to act.
Consider a freeway segment at 55\,mph at 6:45\,AM as morning demand builds: it will either hold near free-flow or collapse to 15\,mph once a downstream bottleneck activates.
During stable conditions real-time observations suffice, but at the edge of breakdown only a forward-looking prediction can distinguish a 4-minute from a 16-minute traversal.
The root cause is a shape mismatch: the true future speed distribution during transitions is bimodal (Figure~\ref{fig:distributions}), but in zero-shot mode all three TSFMs produce effectively unimodal prediction intervals centered between the modes.
The congested regime is itself unimodal (concentrated near 10--20\,mph), so the problem there is merely one of interval width; transitions are fundamentally bimodal, making this a problem of distributional \emph{shape}.

\subsection{Post-Hoc Coverage Correction}

Table~\ref{tab:methods_cov} compares the four post-hoc methods on transition-regime coverage at $h$=12.

\textbf{Global ACI} barely moves the needle: +0.5--1.2\,pp on transition coverage.
This is expected: widening a unimodal interval centered at the wrong location cannot recover the missing mode.

\textbf{Regime-Conditional ACI} does slightly better (+2.3--2.9\,pp on PEMS-BAY) because the transition-specific $\alpha$ adapts more aggressively.
But it still only adjusts width, not shape.

\textbf{BMA} produces the largest improvements, ranging from +2.6\,pp (Moirai, PEMS-BAY) to +16.3\,pp (Chronos-Bolt, PEMS-BAY).
The improvement is largest for Chronos-Bolt, which has the worst native coverage: BMA lifts its PEMS-BAY transition coverage from 54.9\% to 71.2\%.
\textbf{BMA + ACI} adds a further 1--2\,pp on top of BMA by adjusting residual width, achieving the best overall results.
The two methods are complementary: BMA corrects distributional \emph{shape} (injecting the missing mode) while ACI corrects residual \emph{width} (scaling the interval to account for remaining miscoverage).
We use global ACI rather than regime-conditional ACI in the combination because once BMA has corrected the shape mismatch, the residual coverage error is uniform enough across regimes that per-regime width adjustment offers no additional benefit.

\begin{table}[t]
\caption{Transition-regime coverage (\%) at $h$=12, 90\% nominal. The historical conditional baseline (Hist.) achieves high transition coverage by sampling from training data. BMA approaches this level while preserving the TSFM's superior point accuracy.}
\label{tab:methods_cov}
\vskip 0.05in
\centering
\footnotesize
\setlength{\tabcolsep}{2pt}
\begin{tabular}{@{}l l cccccc@{}}
\toprule
& Model & Hist. & Native & G-ACI & R-ACI & BMA & +ACI \\
\midrule
\multirow{3}{*}{\rotatebox[origin=c]{45}{\scriptsize METR-LA}}
& Chr-T5   & \multirow{3}{*}{81.6} & 68.2 & 71.3 & 70.7 & 78.3 & \textbf{81.9} \\
& Chr-Bolt &       & 68.6 & 69.7 & 70.2 & 78.7 & \textbf{80.3} \\
& Moirai   &       & 80.3 & 80.9 & 81.6 & 84.1 & \textbf{84.5} \\
\midrule
\multirow{3}{*}{\rotatebox[origin=c]{45}{\scriptsize P-BAY}}
& Chr-T5   & \multirow{3}{*}{81.7} & 65.0 & 65.5 & 67.9 & 76.7 & \textbf{77.2} \\
& Chr-Bolt &       & 54.9 & 56.1 & 57.5 & 71.2 & \textbf{73.6} \\
& Moirai   &       & 71.6 & 72.1 & 74.3 & 74.2 & \textbf{74.9} \\
\bottomrule
\end{tabular}
\end{table}

The historical conditional baseline (Table~\ref{tab:methods_cov}, ``Hist.'') achieves 81--82\% transition coverage by construction, since it samples directly from the bimodal historical distribution.
But it has poor overall MAE (Table~\ref{tab:regime_mae}), because it is a lookup table, not a forecaster.
TSFMs provide the opposite: good overall accuracy but poor transition coverage.
BMA combines both: it keeps the TSFM's point predictions while borrowing the historical distribution's shape, approaching the historical baseline's transition coverage without sacrificing overall accuracy.

BMA uses per-sensor historical data, just as ACI uses past prediction residuals.
The TSFM itself remains zero-shot; BMA is a post-hoc correction applied to its outputs, analogous to how conformal methods adjust intervals using domain data.
BMA does not degrade coverage in other regimes because the replacement fraction is modulated by the transition probability: in stable free-flow ($P(\text{congested}) \approx 0$), no samples are replaced.
Results are stable across mixing weights ($w \in [0.2, 0.5]$) and degrade only below $w = 0.1$.

\section{Discussion and Limitations}
\label{sec:discussion}

\paragraph{Implications for benchmarks.}
Our results suggest that aggregate evaluation metrics are insufficient for domains with regime switching.
Regime-stratified evaluation is straightforward to implement and could be applied to any domain with distinct operating regimes: electricity spot prices switch between normal and spike regimes \citep{huisman2003regime,energies2025electricity}, wind power near turbine cut-in speeds exhibits similar bimodality, and financial markets alternate between calm and turbulent volatility regimes \citep{Cheng2026,cheng2026volatility}.

\paragraph{BMA and the zero-shot boundary.}
BMA uses per-sensor historical data from the training set, just as conformal methods use past residuals.
The TSFM itself remains zero-shot; BMA is applied to its outputs.
An alternative is to fine-tune the TSFM on traffic data directly, which may teach it to produce bimodal predictions natively.
Whether fine-tuning closes the gap is an open empirical question.
BMA remains relevant when fine-tuning is impractical or when using closed-source models.

\paragraph{Other limitations.}
Our evaluation uses 30 congestion-prone sensors per dataset with univariate forecasts; spatial information from upstream and downstream sensors could improve both accuracy and coverage.
BMA's mixing weight is tuned on 10 held-out windows; a validation-set approach would be more principled.
The interval width increase ($\sim$50--80\%) trades sharpness for coverage; whether this is acceptable depends on the application.

\bibliography{references}
\bibliographystyle{icml2026}

\end{document}